%% file: main.tex
\documentclass[10pt,twocolumn,letterpaper]{article}
\usepackage[numbers,sort,compress]{natbib}

\usepackage{cvpr}
\usepackage{times}
\usepackage{epsfig}
\usepackage{adjustbox}

\usepackage[utf8]{inputenc}

\usepackage{amsmath, amsthm, amssymb}

\usepackage[english]{babel}
\usepackage{bm}

\usepackage[pagebackref=false,breaklinks=true,letterpaper=true,colorlinks,bookmarks=false]{hyperref}

\cvprfinalcopy %

\def\cvprPaperID{****} %

\ifcvprfinal\fi

\title{Human 3D keypoints via spatial uncertainty modeling}
\input{sec/00_authors}

\input{preamble}

\input{math}

\begin{document}
\maketitle
\input{sec/00_abstract}
\input{sec/01_intro}

\input{sec/03_method}

\input{sec/04_results}

\input{sec/05_future}

{
    \small
    \setlength{\bibsep}{0pt}
    \bibliographystyle{abbrvnat}
    \bibliography{macros,main}
}
\end{document}

%% file: sec/00_authors.tex
\author{
Francis Williams$^{1,4}$, \hspace{.5em}
Or Litany$^{2,3}$, \hspace{.5em}
Avneesh Sud$^{1}$, \hspace{.5em}
Kevin Swersky$^{1}$, \hspace{.5em}
Andrea Tagliasacchi$^{1,5}$ \hspace{.5em}
\\[.2in]
$^1$Google Research, \hspace{.5em}
$^2$Stanford University, \hspace{.5em}
$^3$NVIDIA, \hspace{.5em}
$^4$NYU, \hspace{.5em}
$^5$University of Toronto \hspace{.5em}
}

%% file: preamble.tex
\usepackage{overpic}
\usepackage{microtype}
\usepackage{enumitem}
\usepackage{overpic}

\usepackage{color}
\definecolor{turquoise}{cmyk}{0.65,0,0.1,0.3}
\definecolor{purple}{rgb}{0.65,0,0.65}
\definecolor{dark_green}{rgb}{0, 0.5, 0}
\definecolor{orange}{rgb}{0.8, 0.6, 0.2}
\definecolor{red}{rgb}{0.8, 0.2, 0.2}
\definecolor{darkred}{rgb}{0.6, 0.1, 0.05}
\definecolor{blueish}{rgb}{0.0, 0.3, .6}
\definecolor{light_gray}{rgb}{0.7, 0.7, .7}
\definecolor{pink}{rgb}{1, 0, 1}
\definecolor{greyblue}{rgb}{0.25, 0.25, 1}

\renewcommand{\paragraph}[1]{\vspace{0em}\noindent\textbf{#1}.}
\usepackage{blindtext}

\newcommand{\CIRCLE}[1]{\raisebox{.5pt}{\footnotesize \textcircled{\raisebox{-.75pt}{#1}}}}

%% file: math.tex
\DeclareMathOperator{\softmax}{softmax}

\newcommand{\real}{\mathbb{R}}
\newcommand{\R}{\mathbb{R}}

%% file: sec/00_abstract.tex
\begin{abstract}
We introduce a technique for 3D human keypoint estimation that directly models the notion of spatial uncertainty of a keypoint. Our technique employs a principled approach to modelling spatial uncertainty inspired from techniques in robust statistics. Furthermore, our pipeline requires no 3D ground truth labels, relying instead on (possibly noisy) 2D image-level keypoints. Our method achieves near state-of-the-art performance on Human3.6m while being efficient to evaluate and straightforward to implement.
\end{abstract}

%% file: sec/01_intro.tex
\section{Introduction}\label{sec:project_overview}
The ability to identify semantic human keypoints is a classical problem in computer vision that finds many applications in the real world spanning from gaming~\cite{shotton2011real}, athletics~\cite{rhodin2018learning}, robotics~\cite{angjoo}, and has now entered our household in allowing us to control smart devices with our body~(e.g.~Google Nest Hub Max and Facebook Portal).
In the supervised setting, the identification of 2D keypoints is typically considered a ``solved'' problem~\cite{papandreou2018personlab,openpose}, with most of the recent research efforts either targeting 3D keypoint estimation~\cite{angjoo,martinez2017simple,tome2017lifting}, or having moved to the more challenging unsupervised learning setting~\cite{jakab2018unsupervised}.

While the success of 2D keypoint estimation pipelines is largely due to the ease of generating 2D annotations, specifying ground truth 3D keypoints from a \textit{single} image is ill-posed. Hence, researchers have proposed circumventing this issue by leveraging statistical body models~\cite{angjoo}, motion-capture data~\cite{tome2017lifting}, or resorting to multi-camera setups~\cite{simon2017hand,iskakov2019learnable} where 3D is recovered from 2D estimates via \textit{triangulation}.
In this paper, we perform several core technical contributions \emph{to the latter}, towards the general objective of enabling high-quality motion capture to ``in-the-wild'' settings.
In particular, existing multi-camera techniques~\cite{iskakov2019learnable, kocabas2019selfsupervised, pavlakos2017harvesting, rhodin2018unsupervised, rhodin2018learning}
suffer from two significant shortcomings:
\CIRCLE{1} they assume a training set with a sufficiently large set of cameras to minimize self-occlusion and enable accurate extraction of ground truth 3D labels;
\CIRCLE{2} they are not equipped with spatial uncertainty estimates about their prediction.

If we hope to generalize the performance of multi-view setups to computer vision and be robust to outliers in ``in-the-wild'' (i.e. outside capture studio alike~\cite{h36m_pami}), then both these shortcomings ought to be improved:
\CIRCLE{1} our architecture requires only 2D labels which possibly include noise, and
\CIRCLE{2} our model produces an interpretable notion of keypoint uncertainty.
We realize these improvements via a principled approach to modelling uncertainty: representing a keypoint as a parameterized probability distribution in~3D space that can be marginalized onto an image plane to compute a loss in~2D. Our modelling approach is motivated by robust statistics to enable training with large outliers in the labels. Finally, while we use explicit 2D labels for our experiments, we remark that our method could be trained on the output of an ensemble of off-the-shelf 2D keypoint models, leading to a fully unsupervised 3D human pose pipeline.

%% file: sec/03_method.tex
\section{Method}\label{sec:method}
\input{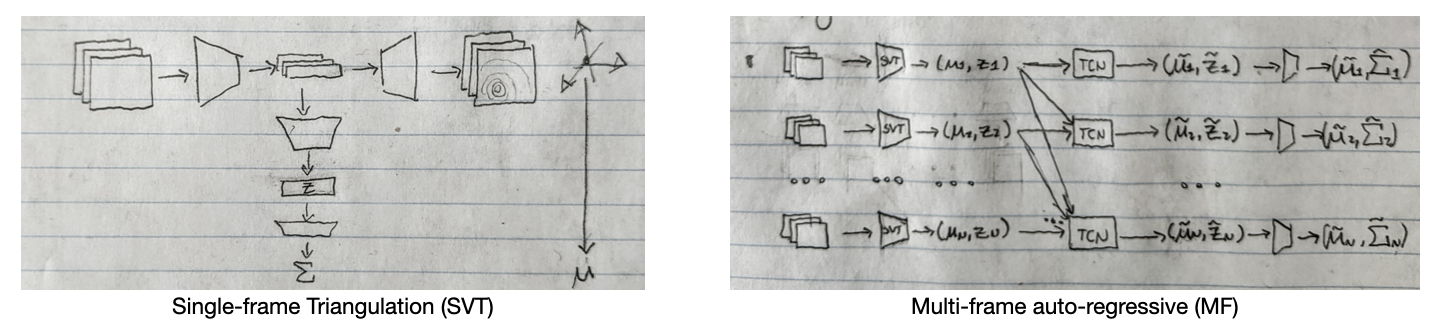}
As input, we are given a multi-view setup consisting of $C$ cameras with known extrinsics and intrinsics $\{\bm P_c \in \real^{3 \times 4}\}$ capturing images $\{\mathcal{I}_c \in \mathbb{R}^{3 \times w \times h}\}$ where $w$ and $h$ are the width and height of the image width. We then seek to predict $K$ spatial distributions (over $\R^3$) corresponding to the $K$ keypoints of a human subject. We assume as a prior that each keypoint follows a multivariate t-distribution with mean $\bm \mu$ and scale matrix $\Sigma$:
\begin{equation}\label{eq:mvtdist}
P(\bm x) \propto [1 + \frac{1}{\nu}(\bm x - \bm \mu) \Sigma^{-1}(\bm x - \bm \mu)]^\frac{-(\nu+3)}{2}    
\end{equation}
where $\nu$ is a hyperparameter which controls the falloff of the t-distribution. A large value of $\nu$, leads to a distribution which is more robust to outliers (due to a faster falloff), but slower to train; we found via a parameter sweep that setting $\nu{=}5$ led to good results. We remark that the covariance of our distribution is a constant multiple of the scale matrix, and therefore refer to $\Sigma$ as the covariance in what follows.
Our choice of parameterization is motivated by three factors:
\CIRCLE{1} it leads to interpretable keypoints, since, similar to a Gaussian, we can view the mean $\bm \mu$ as the most likely spatial location of a keypoint, and the covariance $\Sigma$ as defining a spatial uncertainty;
\CIRCLE{2} the log likelihood loss of a multivariate-t distribution is robust to outliers, due to its rapidly decaying density;
\CIRCLE{3} a multivariate t-distribution can be projected with perspective onto a 2D version of itself, enabling 2D supervision.

Similarly to \textit{Algebraic Triangulation} proposed by~\citet{iskakov2019learnable}, our method consists of a 2D backbone network applied to each view, followed by a differentiable triangulation step.
Unlike \citet{iskakov2019learnable}, our method directly models keypoint uncertainty, outputting a 3D distribution~(as opposed to a point-wise quantity) while requiring only 2D labels which can contain noise; see Figure~\ref{fig:arch} for a schematic of our architecture.

\paragraph{Predicting keypoint distributions}
\label{sec:predicting_dists}
To output a multivariate t-distribution, we must predict its two parameters $\bm \mu$ and $\Sigma$. To predict $\bm \mu$, we follow the Algebraic triangulation method in \cite{iskakov2019learnable}. First, a 2D backbone network accepts the $C$ views of a subject as input and produces $C$ collections of $K$ heatmaps $\{H_{c, k}\}_{c\in [C], k \in [K]}$ for each keypoint in each view. We apply a global softmax activation to each of these heatmaps to convert them to a probability distribution, and take the spatial mean of the image to generate a 2D keypoint prediction
\begin{equation}
    \bm \mu_{c, k}^{\text{2D}} = \sum_{i, j = 0}^{w, h} \left(\begin{matrix}i \\ j\end{matrix}\right)\softmax(H_{c, k})[i, j]
\end{equation} where $H_{c, k}[i, j]$ is the pixel at coordinate $(i, j)$ of the heatmap for keypoint $k$ in view $c$, and $w, h$ are the width and height of the heatmap images.
Again following~\citet{iskakov2019learnable}, we use the architecture from~\citet{xiao2018simple} as our 2D backbone, which is initialized with pretrained weights.

Similarly to~\citet{iskakov2019learnable}, to output 3D mean estimates, we triangulate the 2D predictions by solving a total least squares problem~\cite{hartley2003multiple} which amounts to minimizing a quadratic energy formed by the intersection of points in projective space subject to a unit norm constraint.

Similarly to~\citet{kumar2019uglli}, to predict the covariance~$\Sigma$, we use the bottleneck layer for each image, $z_1, \ldots z_C$, $z_i {\in} \R^{128}$ of the backbone network.
Since the uncertainty about a keypoint should be a function of all views of a person, we aggregate these bottleneck features and predict a lower triangular matrix $L {\in} \R^{3 \times 3}$ as 
\begin{equation}
    L = \rho\bigg(\frac{1}{C}\sum_{i=1}^C \phi(z_i)\bigg)
\end{equation}
where $\rho$ and $\phi$ are both fully connected networks. We apply a shifted ELU activation ($ELU(x) + 1$) to the diagonal entries of $L$ to ensure they are strictly positive and then compute sigma as 
\begin{equation}
    \Sigma = LL^T
\end{equation}
which is positive definite by construction; note we omit the keypoint index $k$ for brevity of notation.

\paragraph{Supervision}
3D keypoint labels are often acquired via some kind of triangulation process which may introduce bias into the labels, which will be learned downstream by the network.
Thus, we propose to learn the triangulation directly from a collection of~(possibly noisy) 2D keypoint labels in each camera view. 
To train our model, we maximize the likelihood of these 2D labels under a projected version of our 3D keypoint distribution.
In contrast to training directly on 3D data, our method aims to produce 3D predictions whose projections agree with 2D keypoint labels under perspective projection. Moreover, training in this way allows our method to naturally consider an ensemble 2D labels (given, for example, by multiple 2D keypoint models or multiple labellers), aggregating these results in a way that is robust to noise and outliers.

\paragraph{Projecting keypoint distributions}
To supervise on 2D labels, we project our keypoint distributions from 3D onto the image planes of each.
Furthermore, we require that the projections of our 3D distributions have an explicit form so we can evaluate a loss in each image.
We first observe that, similar to a Gaussian, for any point $x$ sampled from a multivariate t-distribution \eqref{eq:mvtdist}, its image $x' = Ax + b$ under an affine transformation also follows a multivariate t with parameters $\bm \mu' = A\bm \mu + b$ and $\Sigma' = A \Sigma A^T$. While perspective projection is \emph{not} an affine transformation, we can rely on para-perspective projection \cite{hartley2003multiple}, an affine approximation to perspective to map our distributions from 3D to 2D. We remark that \cite{yamashita20193dgmnet} use a similar projection technique for projecting Gaussians from 3D to 2D. 
Assuming that $x$ and $\bm \mu$ are expressed in the same coordinate system as the camera (where the positive $z$-axis is pointing in the ``look-at'' direction of the camera, and the origin is the camera center), we can write the para-perspective operation as
\begin{equation}\label{eq:para-perspective}
x' = \frac{1}{\|\bm \mu\|} \Pi x, \qquad \Pi = \begin{bmatrix}1 & 0 & 0\\ 0 & 1 & 0\\0 & 0 & 0\end{bmatrix}
\end{equation}
which is equivalent to first projecting the point $x$ onto a plane centered at $\bm \mu$ parallel to the image plane, then scaling the projection of $x$ by the distance between the mean $\bm \mu$ and the camera origin. We remark that because of the assumed camera orientation the norm $\|\bm \mu\|$ is the depth of the keypoint.
Applying the projection \eqref{eq:para-perspective} to each keypoint and each view, we get $C$ multivariate t-distributions on each image plane.
To train, we minimize the negative log likelihood of 2D keypoint labels $k_1, \ldots, k_K$ under these distributions. Taking the negative log of \eqref{eq:mvtdist}, yields the following loss for each predicted keypoint $\bm \mu_i, \Sigma_i$ and 2D label $k_i$ within a given camera view $c$:
\begin{align}\label{eq:lossfun}
    L_c(\bm \mu_k, \Sigma_k) &= \log|\Sigma_k'| + (\nu + 2) \log \| k_i - \bm \mu_k' \|^2_{\Sigma_k'^{-1}}
\end{align}
where we use Mahalanobis norms to short-hand notation, and $\bm \mu'_k$ and $\Sigma_k'$ are the para-perspective projections of $\bm \mu_k$ and $\Sigma_k$ onto the image plane:
\begin{equation}
    \bm \mu'_k = \Pi \frac{\bm \mu_k}{\|\bm \mu_k\|}, \qquad \Sigma'_k = \frac{1}{\|\bm \mu\|^2} \,\Pi \,\Sigma_k \, \Pi^T
\end{equation}

Our final loss function is simply the average of the losses of each keypoint in each camera:
\begin{equation}
    \mathcal{L} = \frac{1}{CK}\sum_{c, k = 1}^{C, K} L_c(\bm \mu_k, \Sigma_k)
\end{equation}

%% file: fig/outline.tex
\begin{figure*}
\centering
\includegraphics[width=\linewidth]{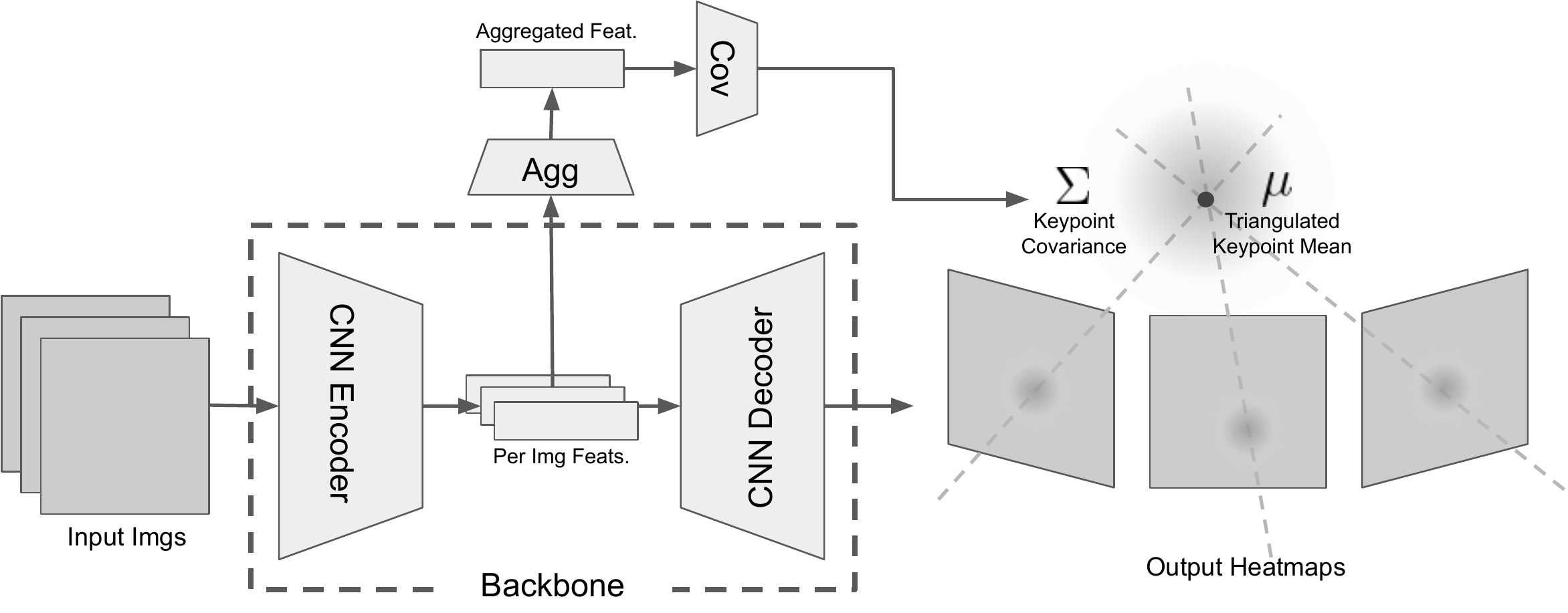}
\caption{\textbf{Model architecture --}
We predict keypoints as probability distributions in 3D parameterized by a mean $\bm \mu$ and covariance $\Sigma$ representing the position and spatial uncertainty of a keypoint.
As input, our model accepts multiple images of a subject taken from multiple~(known) camera views.
We feed these images through a 2D backbone using the architecture from \citet{xiao2018simple} which predicts one heatmap per keypoint within each view.
To predict $\mu$, we triangulate the spatial mean of each heatmap using known camera parameters.
To predict $\Sigma$, we aggregate the bottleneck features of the backbone network, and use a fully connected network to predict a decomposition of $\Sigma$ that is guaranteed to be positive definite.
}
\label{fig:arch}
\end{figure*}

%% file: sec/04_results.tex
\input{fig/results}
\input{tab/fourcam}
\input{tab/twocam}
\input{tab/twocamillposed}

\section{Experiments and results}
We evaluate our method on the widely used Human3.6M dataset \cite{h36m_pami}. Human3.6M consists of video sequences of 7 subjects doing 15 actions taken from 4 cameras. Of these, 2 subjects are reserved for validation and 5 for testing. 
We evaluate our technique against the two state of the art methods proposed in~\citet{iskakov2019learnable}.
While our method is similar in spirit to \textit{algebraic triangulation} (AT) from~\citet{iskakov2019learnable}, we outperform it in our experiments.
In particular, our method is more robust to camera configurations where the triangulation problem is \emph{ill-conditioned}; see Section \ref{sec:ill-cond}.
Our method performs slightly worse than the \textit{volumetric triangulation} (VT) from~\citet{iskakov2019learnable}, however it is much faster to evaluate, conceptually simpler, and requires no data preprocessing to compute bounding boxes, thus making it easier to integrate into practical vision pipelines.
Furthermore, our method provides additional explainability of the results in the form of explicit geometric uncertainty; see Figure~\ref{fig:results} for example predictions from our pipeline.

\subsection{Comparison with 4 cameras -- Table~\ref{tab:4camresults}}
As a baseline, we first compare our method to \citet{iskakov2019learnable} using the full 4 cameras available in the dataset.
As in \citet{iskakov2019learnable}, we report the MPJPE (Mean Per Joint Position Error), which measures the ground L2 distance between the ground truth keypoints relative to the pelvis.
The results are reported in Table~\ref{tab:4camresults}.
Our method slightly outperforms the \textit{algebraic triangulation} method, and performs slightly worse than the \textit{volumetric triangulation} method from~\citet{iskakov2019learnable}.

\subsection{Comparison with 2 cameras -- Table~\ref{tab:2camresults} \& \ref{tab:2camresults-illposed}} 
We now compare our model on the same train/test split as the four camera case, except trained on a subset of 2 cameras and evaluated on 2 different cameras.
In both the train and evaluation setups the cameras are pointing in the same direction, thus introducing possible occluded body parts which could increase error.
We find our method outperforms the algebraic method but underperforms the volumetric method; see Table~\ref{tab:2camresults} for quantitative results.

\paragraph{Robustness Stress Test}\label{sec:ill-cond}
Finally, as a robustness stress test we consider a 2 camera train/test split as above, but where the cameras are antipodal.
This camera configuration leads to many predictions where the triangulation problem is ill-posed, causing large outliers.
We see that in this case, our method is much more robust than algebraic triangulation, but much less robust than the volumetric triangulation; see Table~\ref{tab:2camresults-illposed} for quantitative results.

%% file: fig/results.tex
\begin{figure}
\centering
\includegraphics[width=\columnwidth]{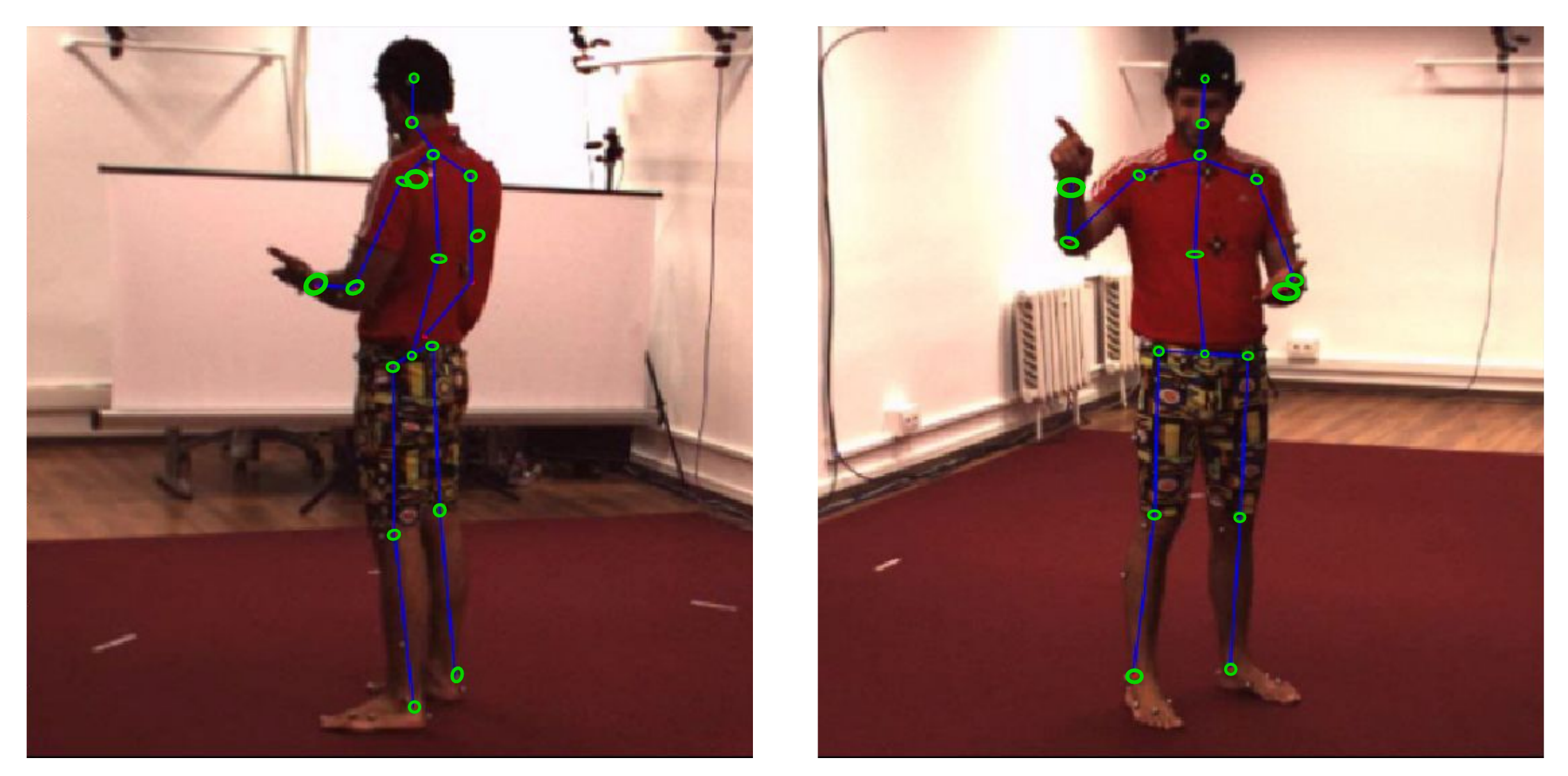}
\includegraphics[width=\columnwidth]{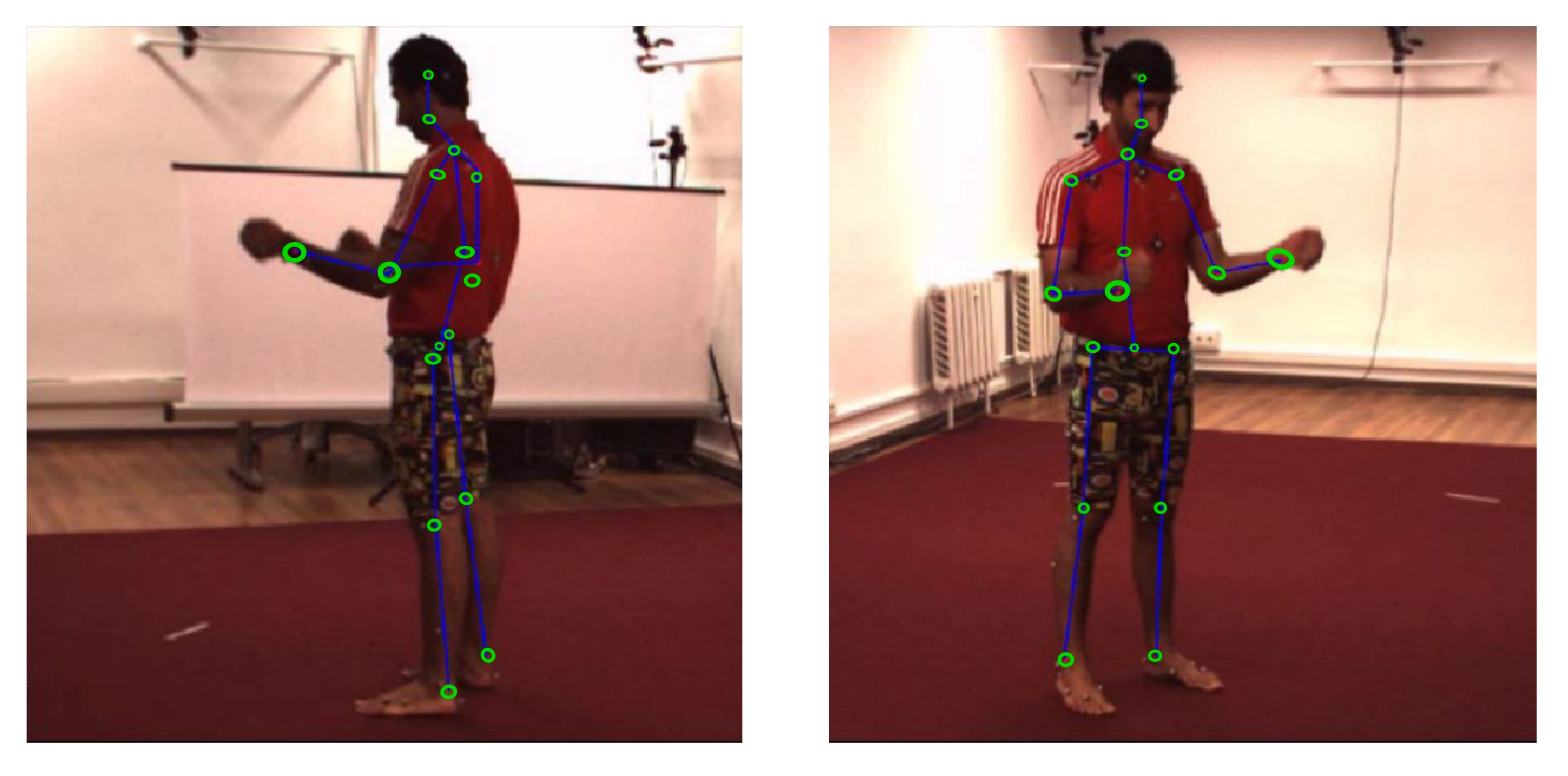}
\caption{Results from the test set of Human3.6M trained on 2 cameras. The green circles represent the 95\% confidence ellipsoid of the predicted keypoint distribution and the red points are the ground truth labels. The wrist keypoints have greater spatial uncertainty since the model makes more error in these areas.
Also note the ground truth keypoints in Human3.6M that are used for training are extremely precise, as they are obtained by tracking a skeleton with a mo-cap system; this justifies the small spatial uncertainty predicted by the model.
}
\label{fig:results}
\end{figure}

%% file: tab/fourcam.tex
\begin{table*}
\centering
\begin{adjustbox}{width=0.9\textwidth}
    \begin{tabular}{c|c|c|c|c|c|c|c|c|c|c|c|c|c|c|c|c}
         Method                              & Avg  & Dir  & Disc & Eat  & Greet & Phone & Pose & Purch & Sit  & SitD & Smoke & Photo & Wait & Walk & WalkD & WalkT \\
         \hline
         AT-Conf \cite{iskakov2019learnable} & 22.6 & 20.4 & 22.6 & 20.5 & 19.7  & 22.1  & 19.5 & 23.0  & 25.8 & 33.0 & 23.0  & 20.6  & 21.6 & 23.7 & 20.7  & 21.3  \\
         VT-Conf \cite{iskakov2019learnable} & 20.8 & 19.9 & 20.0 & 18.9 & 18.5  & 20.5  & 18.4 & 22.1  & 22.5 & 28.7 & 21.2  & 19.4  & 20.8 & 22.1 & 19.7  & 20.2  \\
         Ours                                & 21.6 & 18.3 & 21.6 & 20.0 & 18.8  & 21.1  & 18.9 & 22.4  & 25.2 & 31.1 & 21.9  & 20.6  & 20.7 & 22.7 & 19.6  & 20.1
    \end{tabular}
\end{adjustbox}
\vspace{0.2em}
\caption{MPJPE on all 4 cameras on Human3.6m}
\label{tab:4camresults}
\end{table*}

%% file: tab/twocam.tex
\begin{table*}
\centering
\begin{adjustbox}{width=0.9\textwidth}
    \begin{tabular}{c|c|c|c|c|c|c|c|c|c|c|c|c|c|c|c|c}
         Method                              & Avg   & Dir   & Disc  & Eat  & Greet   & Phone & Pose  & Purch & Sit  & SitD  & Smoke & Photo & Wait & Walk  & WalkD & WalkT \\
         \hline
         AT-Conf \cite{iskakov2019learnable} & 30.9  & 24.3  & 31.2  & 26.7 & 27.7    & 33.2  & 24.2  & 30.3  & 39.0 & 48.4  & 31.8  & 33.0  & 26.6 & 26.5  & 29.7  & 25.8 \\
         VT-Conf \cite{iskakov2019learnable} & 27.8  & 21.4  & 27.2  & 24.3	& 24.1    & 29.1  & 22.2  & 27.7  & 35.3 & 46.4  & 28.9  & 28.3  & 24.1 & 23.2  & 26.9  & 23.2\\
         Ours                                & 30.8  & 24.0  & 30.8  & 26.1 & 28.1    & 33.6  & 23.9  & 30.1  & 39.1 & 49.5	 & 30.9	 & 32.5  & 26.0 & 26.5  & 29.3  & 25.9
    \end{tabular}
\end{adjustbox}
\vspace{0.2em}
\caption{MPJPE on a model trained on 2 cameras (and tested on 2 different cameras) on Human3.6m}
\label{tab:2camresults}
\end{table*}

%% file: tab/twocamillposed.tex
\begin{table*}
\centering
\begin{adjustbox}{width=0.9\textwidth}
    \begin{tabular}{c|c|c|c|c|c|c|c|c|c|c|c|c|c|c|c|c}
         Method                              & Avg   & Dir   & Disc  & Eat  & Greet   & Phone & Pose  & Purch & Sit  & SitD  & Smoke & Photo & Wait & Walk  & WalkD & WalkT \\
         \hline
         AT-Conf \cite{iskakov2019learnable} & 158.4 & 105.4 & 173.4 & 49.9 & 1,029.9 & 67.2  & 55.0  & 68.6  & 70.2 & 505.1 & 61.6  & 76.7  & 75.2 & 40.6  & 61.4  & 50.18 \\
         VT-Conf \cite{iskakov2019learnable} & 39.4  & 34.3  & 36.3  & 37.3 & 43.0    & 40.9  & 31.9  & 41.1  & 44.3 & 54.3  & 40.5  & 39.5  & 41.1 & 32.4  & 37.6  & 34.0 \\
         Ours                                & 90.7  & 118.4 & 108.7 & 56.5 & 154.5   & 69.2  &	86.3  & 66.6  & 75.6 & 183.7 & 69.8  & 96.4  & 90.7 & 45.7  & 79.8  & 59.7
    \end{tabular}
\end{adjustbox}
\vspace{0.2em}
\caption{MPJPE on all model trained on 2 \textit{antipodal} cameras (and tested on 2 \textit{antipodal} cameras) on Human3.6m.}
\label{tab:2camresults-illposed}
\end{table*}

%% file: sec/05_future.tex
\section{Conclusions and Future Work}
In this technical report, we presented a technique for learning 3D keypoints from image (possibly noisy) image labels. Unlike prior work, our method requires only 2D supervision and produces outputs which are equipped with an interpretable notion of spatial uncertainty. Our method achieves near state-of-the-art results on a standard benchmark while remaining extremely simple to implement and fast to evaluate.
We believe that such a method paves the way for important future work in the problem of human keypoint detection.

\paragraph{Fully Unsupervised Keypoint Detection}
We remark that many existing methods exist in the literature for image level keypoint estimation (\eg \cite{tompson2015efficient, Toshev_2014_CVPR, wei2016convolutional}). Since our pipeline requires only noisy 2D annotations, we could leverage an ensemble of 2D predictions from existing models to generate labels on the fly. Our model, would not only learn to predict the mean of this ensemble, but also the covariance, providing an explicit measurement for the consensus between labellers.
Furthermore, the backbone network in our model is, in general, not pretrained on the same dataset as it is evaluated on (for example, our backbone was pretrained on COCO \cite{lin2014microsoft}).
Thus, an ensemble prediction would be fully unsupervised, and could be used, for example, to construct very large datasets using only a few calibrated cameras.

\paragraph{Temporal Predictions}
Spatial uncertainty is a core requirement for many temporal models such as Kalman~Filters.
A natural extension of our pipeline, is to aggregate keypoint predictions over multiple frames to predict the next frame in a way that minimizes uncertainty. Furthermore, leveraging the ensemble predictions described in the previous paragraph, our pipeline could be extended to perform fully unsupervised spatio-temporal keypoint prediction.